\documentclass[letterpaper, 10 pt, conference]{ieeeconf}  

\IEEEoverridecommandlockouts                              

\newcommand{\seq}[3]{\{\mathbf{#1}_{t}\}_{#2}^{#3}}
\newcommand{\states}[3]{\mathbf{#1}_{#2}^{#3}}

\newcommand{\R}{\mathbb{R}}

\newif\ifshowcomments
\showcommentsfalse

\ifshowcomments
  \newcommand{\RAM}[1]{\textbf{\color{red} RAM: #1}}
\else
  \newcommand{\RAM}[1]{}
\fi

\usepackage{amsmath} 
\usepackage{amssymb}
\usepackage{color}
\usepackage{graphicx}
\usepackage{booktabs}
\usepackage{multirow}
\usepackage[size=footnotesize]{caption}
\usepackage{float}
\usepackage{makecell}
\usepackage{hyperref}
\usepackage{algorithm}
\usepackage{algorithmic}
\usepackage{cite}

\usepackage[table]{xcolor}
\definecolor{best}{RGB}{255,170,190}     
\definecolor{second}{RGB}{255,225,160}   

\title{\LARGE \bf
TrackDeform3D: Markerless and Autonomous 3D Keypoint Tracking and Dataset Collection for Deformable Objects
}

\author{
Yeheng Zong$^{1*}$, Yizhou Chen$^{1*}$, Alexander Bowler$^{1}$, Chia-Tung Yang$^{1}$, Ram Vasudevan$^{1}$%
\thanks{$^*$These authors contributed equally.}%
\thanks{$^1$The authors are with the Department of Robotics at the University of Michigan, 2505 Hayward Drive, Ann Arbor, USA \textless yehengz, yizhouch,ramv \textgreater@umich.edu}%
}

\begin{document}

\maketitle
\thispagestyle{empty}
\pagestyle{empty}

\begin{abstract}
Structured 3D representations such as keypoints and meshes offer compact, expressive descriptions of deformable objects, jointly capturing geometric and topological information useful for downstream tasks such as dynamics modeling and motion planning.
However, robustly extracting such representations remains challenging, as current perception methods struggle to handle complex deformations.
Moreover, large-scale 3D data collection remains a bottleneck: existing approaches either require prohibitive data collection efforts, such as labor-intensive annotation or expensive motion capture setups, or rely on simplifying assumptions that break down in unstructured environments.
As a result, large-scale 3D datasets and benchmarks for deformable objects remain scarce.
To address these challenges, this paper presents an affordable and autonomous framework for collecting 3D datasets of deformable objects using only RGB-D cameras. 
The proposed method identifies 3D keypoints and robustly tracks their trajectories, incorporating motion consistency constraints to produce temporally smooth and geometrically coherent data. 
TrackDeform3D is evaluated against several state-of-the-art tracking methods across diverse object categories and demonstrates consistent improvements in both geometric and tracking accuracy.
Using this framework, this paper presents a high-quality, large-scale dataset consisting of 6 deformable objects, totaling 110 minutes of trajectory data. 
The \href{https://roahmlab.github.io/trackDeform3D-core-tracking/}{project website} provides access to the dataset, and the \href{https://github.com/roahmlab/trackDeform3D-core-tracking}{code repository} provides detailed implementations.

\end{abstract}

\section{Introduction}
\label{sec:intro}

Deformable objects such as wire harnesses or fabrics are ubiquitous in homes and factories, yet remain challenging for robots to manipulate.
~\cite{deformsurvey1, deformsurvey2}. 
Because these objects possess infinitely many degrees of freedom, directly modeling their full state is intractable.
Researchers therefore often approximate their state with structured 3D representations such as keypoints or meshes, which provide a compact yet informative description of their geometry and topology~\cite{cotracker1, deformdatamocap1, deformdatamocap2, pointcloud1, pointcloud3, pointcloud4, pointcloud5, pointcloud6, DEFORM, DEFT}.
Large-scale, high-quality keypoint and mesh trajectory datasets of deformable objects could significantly advance robotic manipulation.

Scalable data collection demands platforms that are easy to deploy, robust in operation, and require minimal human involvement. 
High data quality further requires correct topology as well as \textit{consistent indexing}, meaning the same keypoint index always refers to the same physical location on the object across all frames and sessions.
High data quality also demands two forms of geometric consistency:
\textit{Initialization consistency} requires that keypoints are placed at the same locations across different collection sessions, minimizing misalignment across data collection sessions.
\textit{Trajectory consistency} requires that distances between neighboring keypoints remain stable throughout a collection episode, preventing oscillating noise and sudden jumps that would corrupt downstream tasks such as computing stretching or bending forces in physics simulation~\cite{clothsimulation}.
Satisfying all of these requirements simultaneously while collecting data at scale remains an open challenge.

\begin{figure}[t]
    \centering
    \includegraphics[width=\linewidth]{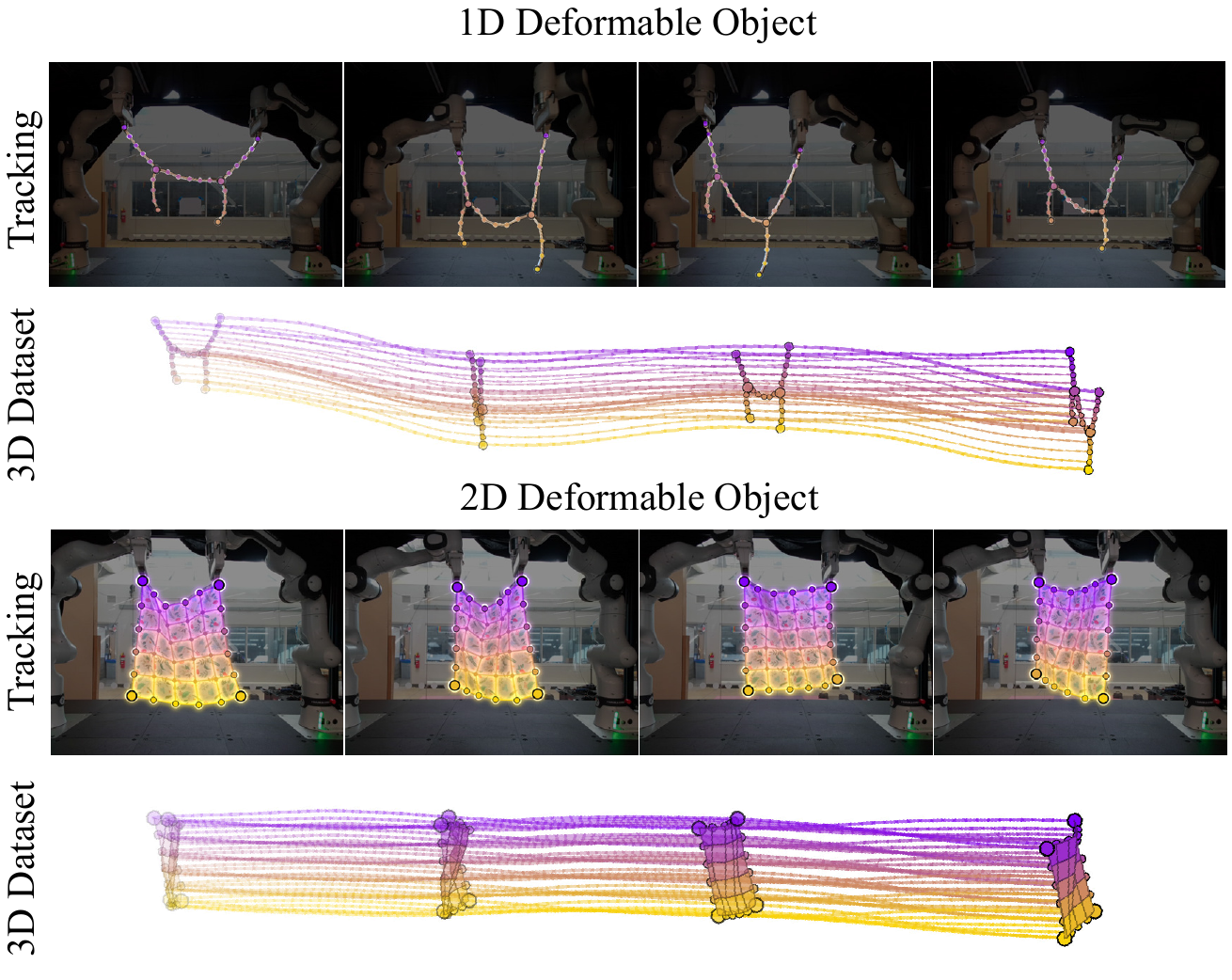}
    \caption{This paper presents TrackDeform3D, a system for keypoint initialization, tracking, and dataset generation for 1D and 2D deformable objects during dual-arm robotic manipulation. 
    For each object category, the figure illustrates tracked keypoint trajectories during manipulation and the resulting 3D trajectory data. 
    Color encodes spatial position across keypoints, while transparency encodes temporal progression, with earlier frames appearing more transparent and later frames more opaque.}
    \label{fig:titlefigure}
    \vspace{-6mm}
\end{figure}

Among existing approaches, marker-assisted motion capture~\cite{deformdatamocap1, deformdatamocap2, DEFORM, DEFT} and volumetric reconstruction systems~\cite{deformvolumetric1, deformvolumetric2} can provide accurate 3D trajectories and meshes. 
However, marker-assisted motion capture requires expensive infrastructure, along with significant human effort for pre- and post-processing.
Volumetric reconstruction methods, meanwhile, are designed for 3D deformable objects such as foam or plasticine and do not generalize to 1D or 2D objects such as wires and cloth \cite{deformvolumetric1, deformvolumetric2}.
Point-cloud-based tracking methods~\cite{pointcloud1, pointcloud3, pointcloud4, pointcloud5, pointcloud6} offer a lightweight alternative, operating with a single RGB-D camera.
However, these methods depend on accurate object segmentation, often relying on simple color filters, and most address neither initialization consistency nor trajectory consistency.

To address these challenges, we propose \textit{TrackDeform3D}, a data collection pipeline for robust 3D keypoint tracking of deformable objects. 
Our contributions are two-fold: 
First, an affordable tracking system that enforces geometric consistency to robustly initialize and track 3D keypoints across multiple collection sessions.
Second, a large-scale, high-quality 3D dataset of 1D and 2D deformable objects comprising 6 objects and totaling 110 minutes of trajectory data for training and benchmarking deformable object models.

\section{Related Work}
Prior work on deformable object tracking and data collection can be broadly categorized into marker-assisted tracking, RGB-D based tracking, and learning-based vision methods, each with limitations that motivate our approach.

\subsection{Marker-assisted Tracking Methods}
MoCap systems~\cite{deformdatamocap1, DEFORM, DEFT} and invisible-marker systems~\cite{deformdatamocap2} can achieve submillimeter accuracy at high temporal resolution, and naturally maintain geometric consistency through physically attached markers.
However, these systems are sensitive to lighting variations and sensor noise, often resulting in missing or corrupted observations.
Recorded data also frequently require manual reindexing to account for changes in object topology and orientation, leading to labor-intensive post-processing.
Finally, these methods rely on costly multi-camera setups, limiting their scalability and deployability in unstructured environments.

\subsection{RGB-D Based Methods}
RGB-D based approaches offer a simpler and more affordable alternative for tracking deformable objects.
Volumetric reconstruction pipelines~\cite{deformvolumetric1, deformvolumetric2} reconstruct 3D meshes of deformable objects but do not generalize to flexible 1D and 2D objects such as wires and cloth undergoing diverse and dynamic motions.
Within the robotics community, several works~\cite{pointcloud1, pointcloud3, pointcloud4, pointcloud5, pointcloud6} have explored point-cloud-based deformable object tracking.
While these methods ensure that extracted keypoints remain on the object surface, most do not enforce trajectory consistency, and none address initialization consistency.
The exception is~\cite{pointcloud5}, which enforces trajectory consistency, but as we show in our experiments, it loses track during dynamic manipulation.
Beyond consistency, these methods assume deformable objects can be easily segmented by color or background contrast, requiring carefully controlled setups, and rely on predefined object topologies rather than inferring them automatically.

The computer vision community has made significant progress in segmentation~\cite{sam1, sam2, samDLO}, visual tracking, and 3D reconstruction, yet directly applying these tools to deformable object data collection presents its own challenges.
Segmentation approaches~\cite{sam1, sam2, samDLO} can isolate deformable objects and lift them into 3D using depth information, but require manual prompting in cluttered scenes and are computationally expensive, taking up to 3 seconds per frame~\cite{samDLO}.
Keypoint-tracking methods~\cite{cotracker1, cotracker3, TAPIP3D, spatialtrackerv2} can track deformable objects in real time but require manual keypoint selection, which is labor-intensive and introduces initialization inconsistency across sequences.
Furthermore, both 2D and 3D tracking methods~\cite{cotracker1, cotracker3, TAPIP3D, spatialtrackerv2} frequently lose track when object and background colors are similar.
These limitations propagate as noise in downstream tasks such as neural network training or system identification, and none of these methods address the geometric and topological consistency required for scalable, autonomous data collection.

\subsection{Deformable Object Dataset}
Few publicly available datasets exist for deformable object manipulation~\cite{deformvolumetric1, deformdatamocap1, deformdatamocap2}.
PokeFlex focuses exclusively on volumetric objects with simple poking interactions ~\cite{deformvolumetric1}.
\cite{deformdatamocap1} is limited to cloth with a small number of motion sequences.
\cite{deformdatamocap2} offers the greatest diversity, but covers only four objects totaling approximately 20 minutes of data.
Using TrackDeform3D, we construct a dataset consisting of 6 deformable objects and 110 minutes of trajectory data, addressing the scale and diversity gaps left by prior work.

\section{Methodology}
\begin{figure*}[t]
    \centering
    \includegraphics[width=\linewidth]{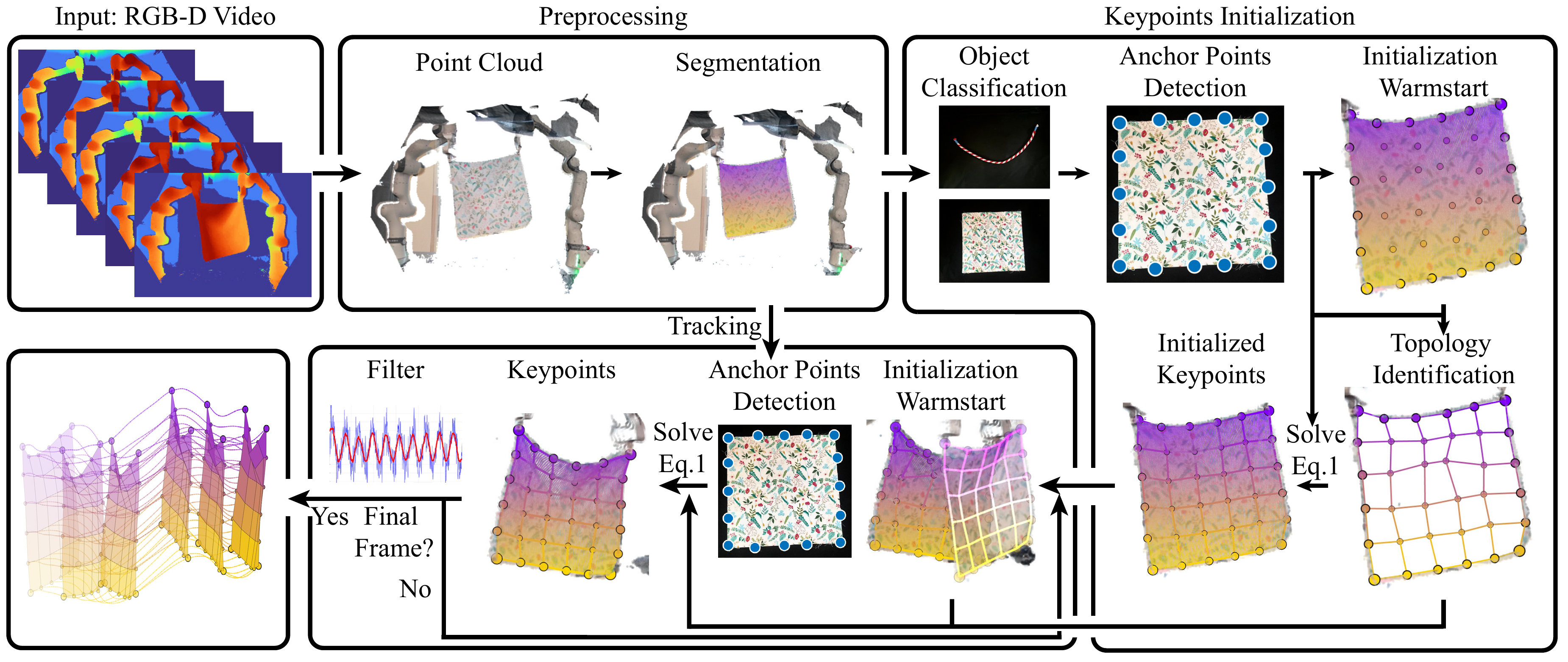}
    \caption{
    Overview of TrackDeform3D.
    Given an RGB-D video, TrackDeform3D first lifts depth images to point clouds and segments the deformable object via point cloud differencing (\S\ref{sec:Segmentation}).
From the first segmented frame, the object type is classified and 3D keypoints are initialized by detecting anchor points, generating warm-start positions, inferring object topology, and solving a constrained geometric optimization (\S\ref{sec:problem_formulation}, \S\ref{sec:initialization}).
During tracking, anchor points are re-detected at each frame and the same optimization is solved recursively, warm-started from the previous solution.
A temporal moving-average filter is applied to suppress high-frequency jitter, producing smooth and temporally consistent 3D keypoint trajectories (\S\ref{sec:tracking}).
}
    \vspace{-2mm}
    \label{fig:algorithmoverview}
\end{figure*}

This section presents TrackDeform3D, our pipeline for robust 3D keypoint tracking of deformable objects.
We begin by describing the experimental setup and problem formulation, then introduce a unified optimization framework that governs both keypoint initialization and tracking.
The initialization stage segments the deformable object and computes an initial keypoint configuration.
The tracking stage then propagates these keypoints throughout the manipulation sequence, producing smooth and temporally consistent trajectories.
An overview of the pipeline is shown in Fig.~\ref{fig:algorithmoverview}.

\subsection{Setup and Assumptions}
\label{sec:setup}

A single RGB-D camera is positioned in front of the data collection station,  where two robotic arms grasp and manipulate the deformable object.
The camera is calibrated with respect to the world frame via hand-eye  calibration, and all quantities discussed hereafter are expressed in this frame.
The camera records a depth video sequence $\seq{D}{t=0}{T} \in \mathbb{R}^{(T+1) \times H \times W \times 1}$  from time $0$ to $T$, where $H$ and $W$ denote the image height and width,  respectively.
The goal is to extract a sequence of 3D keypoints $\seq{X}{t=1}{T} \in \mathbb{R}^{T \times N \times 3}$ from the recorded video, where $N$ is the number of keypoints on the object and each keypoint is represented by its 3D coordinates $(x, y, z)$ in the world frame.

We make three assumptions.
First, the deformable object remains unoccluded throughout manipulation, as data collection requires direct observation of the ground truth object state~\cite{deformdatamocap1, deformdatamocap2, DEFORM, DEFT, deformvolumetric1, deformvolumetric2}.
Second, two robotic arms are available and provide control inputs  $\seq{u}{t=0}{T} \in \mathbb{R}^{(T+1) \times 12}$ for downstream training.
This assumption can be relaxed for human demonstrations by substituting hand pose estimates in place of robot arm inputs~\cite{inhandNN1, inhandNN2}.
Third, the camera intrinsic and extrinsic calibration matrices are known, enabling depth images $\seq{D}{t=0}{T}$ to be lifted to point clouds $\seq{P}{t=0}{T} \in \mathbb{R}^{(T+1) \times H \times W \times 3}$.

\subsection{Problem Formulation}
\label{sec:problem_formulation}

At each time step $t$, we seek to compute keypoint positions $\states{X}{t}{} \!\in\! \R^{N \times 3}$ distributed along the deformable object.
Because the camera observes the full scene, we first segment the object from  the background to obtain a point cloud $\states{P}{t}{seg}$ that constrains keypoints to lie on the object surface.
A natural approach is then to minimize an energy that encourages uniform spacing among neighboring keypoints until an equilibrium is reached.
However, without boundary constraints, such equilibria are sensitive to initialization and can vary across collection sessions, violating initialization consistency.
To address this, we introduce a set of anchor points that serve a dual purpose: 1) they correspond to geometrically significant features relevant to object dynamics (e.g., leaf nodes and junctions for 1D objects, or contour points for 2D objects), which are important landmarks to capture during data collection; and 2) they remain fixed during optimization, constraining the solution space and stabilizing the keypoint configuration across sessions.

Given the point cloud segmentation, the topology of an object, and a set of anchor points, we formulate keypoint computation as the following constrained optimization problem:
\begin{equation}
\label{eq:init_opt}
\begin{aligned}
\arg\min_{\states{X}{t}{}} \;
 & \sum_{(i,j) \in \mathcal{E}} \left( \| \states{x}{t,i}{} - 
 \states{x}{t,j}{} \| - d_{ij} \right)^2 \\
\text{s.t.} \quad 
& \states{x}{t,i}{} \in \states{P}{t}{seg}, \\
& \states{x}{t,k}{} = \states{a}{t,k}{}, \quad \forall \, k \in \mathcal{A},
\end{aligned}
\end{equation}
where $\states{x}{t,i}{}$ denotes the position of the $i$-th keypoint at time $t$, $\mathcal{E}$ is the set of edges defined by the object topology, and $d_{ij}$ is the target edge length between adjacent keypoints $i$ and $j$.
The anchor set $\mathcal{A} \subseteq \{1, \ldots, N\}$ indexes the geometrically significant keypoints, each fixed to a detected position $\states{a}{t,k}{} \in \R^3$.
The edge-length term encourages consistent spacing between adjacent keypoints, the point cloud constraint ensures fidelity to the observed geometry, and the anchor constraints regularize the solution so that it remains well-posed and repeatable across sessions.

Solving~\eqref{eq:init_opt} requires five components: 
1) segmentation of the deformable object point cloud $\states{P}{t}{seg}$,
2) anchor positions $\states{a}{t,k}{}$ and the anchor set $\mathcal{A}$,
3) a topology $\mathcal{E}$ defining connectivity among all $N$ keypoints,
4) warm-start positions for the non-anchor keypoints to improve convergence, and
5) a stable and efficient solver.
We use this formulation for both initialization ($t{=}1$, Section~\ref{sec:initialization}) and tracking ($t \ge 2$, Section~\ref{sec:tracking}), and describe how each component is obtained in the sections that follow.

\subsection{Deformable Object Segmentation}
\label{sec:Segmentation}

The first step is to segment the deformable object from the background.
In unstructured environments, deformable objects often occupy only a small portion of the scene, making robust unsupervised segmentation challenging.
While segmentation networks such as SAM~\cite{sam1, sam2, samDLO} can address this, they require manual prompts to identify the region of interest, which becomes labor-intensive at the scale of large dataset collection.
We therefore propose an autonomous segmentation procedure based on point cloud differencing.

We begin by extracting a point cloud $\states{P}{0}{}$ from the initial depth image $\states{D}{0}{}$, then command the robotic arms to move to a different configuration and capture a second point cloud $\states{P}{1}{}$.
Computing the difference $\states{P}{1}{} - \states{P}{0}{}$ filters out the majority of static background points, leaving three components: robot arm points, deformable object points from both frames, and residual depth sensor noise.
Because the deformable object appears in both frames at different configurations, the differencing yields positive depth values for the object at $t{=}1$ and negative depth values at $t{=}0$, allowing us to isolate the object point cloud $\states{P}{1}{}$.
To remove robot arm points, we reconstruct the arms in simulation using the known joint configurations and camera calibration, and filter out points within proximity of the simulated arm point cloud.
Finally, DBSCAN~\cite{dbscan} is applied to cluster the remaining points.
Because the deformable object constitutes the largest coherent cluster, we select the cluster with the most points, yielding the segmented point cloud $\states{P}{1}{seg}$ for the subsequent stage.

\subsection{Keypoint Initialization}
\label{sec:initialization}
Given the segmented point cloud $\states{P}{1}{seg}$, we compute the initial  keypoints $\states{X}{1}{} \!\in\! \R^{N \times 3}$ by solving~\eqref{eq:init_opt} at $t{=}1$.
The components of~\eqref{eq:init_opt} are determined separately for 1D and 2D deformable objects.
We begin by classifying the object type, then determine the anchor positions, warm-start positions, and topology for each case.
The solver is shared across both topologies.

\subsubsection{\textbf{Object Classification}}
To minimize human involvement, we automatically classify whether the deformable object is 1D or 2D by evaluating local geometry on $\states{P}{1}{seg}$.
We randomly sample seed points $\mathbf{p} \in \states{P}{1}{seg}$ and collect their neighborhoods $\mathcal{N}(\mathbf{p})$ within radius $r$.
Let $\lambda_1 \ge \lambda_2 \ge \lambda_3$ be the eigenvalues of the covariance matrix of $\mathcal{N}(\mathbf{p})$.
A 1D structure is locally line-like, yielding $\lambda_1 \gg \lambda_2 \approx \lambda_3$, whereas a 2D surface is locally planar, yielding $\lambda_1 \approx \lambda_2 \gg \lambda_3$\cite{eigenvalueclassification}.
We compute the ratios $\lambda_2 / \lambda_1$ and $\lambda_3 / \lambda_2$ over the sampled seeds and classify the object type using a fixed threshold.

\subsubsection{\textbf{1D Objects}}
\begin{figure}[t]
    \centering
    \includegraphics[width=\linewidth]{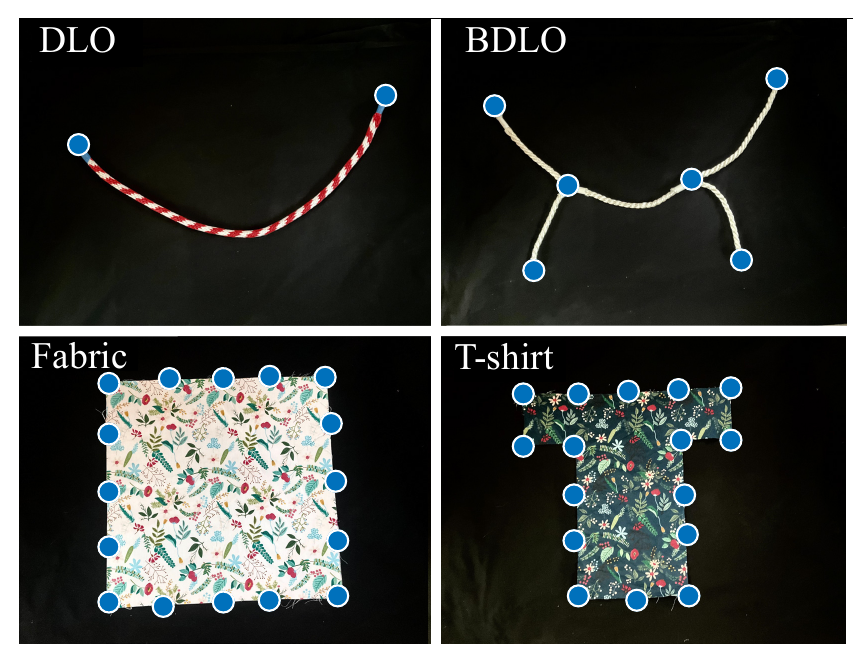}
    \caption{An illustration of anchor point placement on different deformable object categories: DLO, BDLO, fabric, and T-shirt.}
    \label{fig:anchorpoints}
    \vspace{-5mm}
\end{figure}

\textit{Anchor Point Detection.}
For 1D objects such as Deformable Linear Objects (DLOs), anchor points correspond to leaf nodes and junction points.
Leaf nodes mark where the object terminates, which is essential for accurately representing its geometry.
For objects with more complex topology, such as Branched DLOs (BDLOs), junction points are additionally detected to capture the branching structure and provide boundaries for distributing keypoints along each branch.
 An illustration of junction points can be found in Fig.~\ref{fig:anchorpoints}.

To detect these anchor points, we project $\states{P}{1}{seg}$ onto the image plane to obtain a 2D binary mask, then apply skeletonization~\cite{topologyDLO} to extract the medial axis.
A Minimum Spanning Tree (MST) is computed over the skeleton pixels using Euclidean distances as edge weights.
For each skeleton pixel, we compute its degree $s$, defined as the number of incident edges in the MST, and classify it as a \emph{leaf node} if $s \!=\! 1$ or a \emph{junction} if $s \!\ge\! 3$.
The detected points are then lifted back to 3D using the depth image, yielding the anchor positions $\{\mathbf{a}_k\}_{k \in \mathcal{A}}$.

\noindent \textit{Warm Start.}
Non-anchor keypoints are initialized using Farthest Point Sampling (FPS)~\cite{FPS} on $\states{P}{1}{seg}$, seeded from the anchor points $\{\mathbf{a}_k\}_{k \in \mathcal{A}}$.
FPS greedily selects points that maximize the minimum distance to all previously selected points, and starting from the anchors, selects the $N - N_a$ remaining keypoints from $\states{P}{1}{seg}$ to form the warm start for~\eqref{eq:init_opt}.

\noindent \textit{Topology.}
The topology of a 1D object is a chain, or a tree of chains for BDLOs.
Keypoints are ordered by finding the chain sequence that minimizes the total sum of squared edge lengths, and $\mathcal{E}$ is defined as the edges connecting consecutive keypoints in this ordering.

\subsubsection{\textbf{2D Objects (Cloth-like)}}
\textit{Anchor Point Detection.}
For cloth-like objects, anchor points are distributed along the object contour, which captures the overall geometric shape of the cloth.
Dense contour points are first extracted from the segmented point cloud using OpenCV~\cite{opencv_library}, then FPS is applied to distribute anchor points evenly along the contour, yielding $\{\mathbf{a}_k\}_{k \in \mathcal{A}}$.

\noindent \textit{Warm Start and Topology.}
Consistent initialization and topology recovery for 2D objects are more challenging than in the 1D case due to increased geometric complexity and additional degrees of freedom.
Prior work addresses this through template matching against pre-defined cloth meshes~\cite{clothtemplate}.
Here, we exploit the structure of rectangular fabrics, which can be discretized into a regular grid defined by the extracted contours.
Interior keypoints are generated via bilinear interpolation and indexed according to the resulting grid topology.
This approach generalizes to non-rectangular 2D objects by replacing the regular grid with a topology-aware mesh defined by the object boundary

\subsubsection{\textbf{Solve~\eqref{eq:init_opt}}}
For notational brevity, we omit the time subscript $t$ in the following.
We solve~\eqref{eq:init_opt} using a Gauss-Seidel-style iterative projection procedure summarized in Algorithm~\ref{alg:keypoint_init}.
At each iteration, the edge length projection enforces target distances $d_{ij}$ edge by edge, the point cloud projection maps non-anchor keypoints to their nearest neighbor on $\states{P}{}{seg}$, and the anchor projection resets anchor positions to their fixed values $\mathbf{a}_k$.
Together, these projections produce a consistent keypoint configuration.
The initial configuration $\states{X}{1}{}$ is obtained by applying Algorithm~\ref{alg:keypoint_init} to the first segmented point cloud frame.

\begin{algorithm}[t]
\caption{Solve~\eqref{eq:init_opt} via Gauss-Seidel Method}
\label{alg:keypoint_init}
\begin{algorithmic}[1]
\REQUIRE $\states{P}{}{seg}$, $\mathcal{E}$, $\mathcal{A}$, $\{\mathbf{a}_k\}_{k \in \mathcal{A}}$, $\{d_{ij}\}_{(i,j) \in \mathcal{E}}$, $\#$ of iterations $M$
\STATE Initialize $\mathbf{X}$; set $\mathbf{x}_k \leftarrow \mathbf{a}_k, \forall k \in \mathcal{A}$
\FOR{$m = 1$ to $M$}
    \FOR{$(i,j) \in \mathcal{E}$}
    \STATE Project $\mathbf{x}_i, \mathbf{x}_j$ s.t.\ $\|\mathbf{x}_i {-} \mathbf{x}_j\| {=} d_{ij}$ as in \cite[(8)]{projectivedynamics}
    \ENDFOR 
    \STATE $\mathbf{x}_l \leftarrow \arg\min_{\mathbf{p} \in \states{P}{}{seg}} \| \mathbf{x}_l - \mathbf{p} \|, \; \forall l \notin \mathcal{A}$ 
    \STATE $\mathbf{x}_k \leftarrow \mathbf{a}_k, \; \forall k \in \mathcal{A}$ 
\ENDFOR
\RETURN $\mathbf{X}$
\end{algorithmic}
\end{algorithm}

\subsection{Tracking}
\label{sec:tracking}
Given the initialized keypoints $\states{X}{1}{}$, we track them across subsequent frames $t = 2, \ldots, T$ by solving~\eqref{eq:init_opt} at each time step.
The topology $\mathcal{E}$ is fixed after initialization, while anchor positions $\{\states{a}{t,k}{}\}_{k \in \mathcal{A}}$ are re-detected independently at each frame using the procedure described in Section~\ref{sec:initialization}.
At time $t$, the optimization is warm-started from $\states{X}{t-1}{}$ and solved against the segmented point cloud $\states{P}{t}{seg}$, enforcing edge-length and point cloud consistency at each step.

As a post-processing step, we apply a temporal moving average filter to reduce high-frequency jitter.
The position of each keypoint is replaced by the average of positions within a local symmetric temporal window, yielding smooth and temporally consistent trajectories while preserving overall motion.
The full tracking procedure is summarized in Algorithm~\ref{alg:tracking}.

\begin{algorithm}[t]
\caption{Keypoint Tracking}
\label{alg:tracking}
\begin{algorithmic}[1]
\REQUIRE Initial keypoints $\states{X}{1}{}$, segmented point clouds $\{\states{P}{t}{seg}\}_{t=2}^{T}$, topology $\mathcal{E}$, edge lengths $\{d_{ij}\}_{(i,j)\in\mathcal{E}}$
\FOR{$t = 2$ to $T$}
    \STATE $\{\states{a}{t,k}{}\}_{k \in \mathcal{A}} \leftarrow \text{Detect Anchors}(\states{P}{t}{seg})$
    \STATE $\states{X}{t}{} \leftarrow \text{Solve}~\eqref{eq:init_opt}$ with warm start $\states{X}{t-1}{}$ 
\ENDFOR
\STATE $\{\states{X}{t}{}\}_{t=1}^{T} \leftarrow \text{Temporal Average}(\{\states{X}{t}{}\}_{t=1}^{T})$ 
\RETURN $\{\states{X}{t}{}\}_{t=1}^{T}$
\end{algorithmic}
\end{algorithm}
\section{Experiment}
\label{sec:experiment}

This section presents experimental validation of TrackDeform3D as a robust and scalable pipeline for deformable object data collection.
We evaluate tracking performance against baseline methods, ablate the contribution of each component in~\eqref{eq:init_opt}, and compare our collected dataset against existing benchmarks.

\subsection{Experiment Details}
\subsubsection{Hardware Setup}
\begin{figure}[t]
    \centering
    \includegraphics[width=\linewidth]{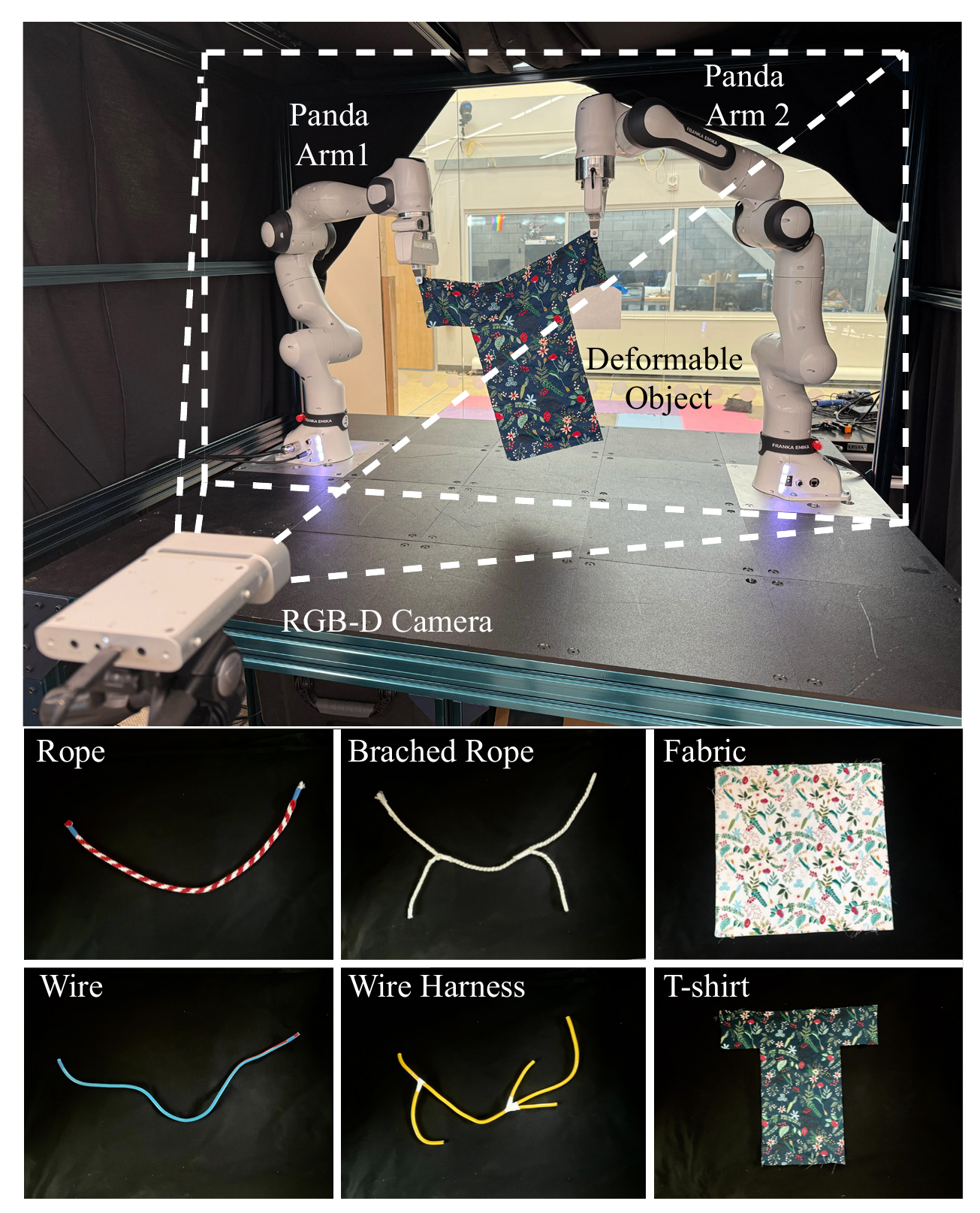}
    \caption{\textbf{Top.} Experimental setup. \textbf{Bottom.} Constructed deformable objects..}
    \label{fig:experiment_setup}
    \vspace{-5mm}
\end{figure}

The data collection station consists of two Franka Emika Panda robotic arms and an RGB-D camera.
The two arms grasp and manipulate deformable objects within a shared workspace, while the camera captures synchronized color and depth streams at 30\,Hz at a resolution of $720 \times 1080$.
The camera is mounted in front of the workspace and calibrated with respect to both robot base frames via hand-eye calibration.
An overview of the setup is shown in the upper portion of Fig.~\ref{fig:experiment_setup}.

\subsubsection{Dataset Collection Preparation}
We assemble a diverse dataset spanning both 1D and 2D deformable objects.
The 1D category includes DLOs such as rope and wire, as well as BDLOs such as branched rope and wire harnesses.
The 2D category includes rectangular fabric and T-shirt.
Examples are shown in the lower portion of Fig.~\ref{fig:experiment_setup}.

Within each category, objects are varied along several dimensions to ensure diversity and evaluate the robustness of TrackDeform3D.
\textit{Stiffness}: 1D objects range from flexible rope to relatively stiff 
wire with plasticity.
\textit{Appearance}: objects span uniform color (wire, wire harness, branched 
rope), patterned color (rope, fabric), and random patterns (cloth).
\textit{Contact}: sequences include both contact and non-contact configurations, where the object contacts the workstation table.
Upon release, these will be split into separate subsets, with the workstation table mesh provided for contact-aware benchmarking.

All data is collected using pre-recorded trajectories executed by the two Panda arms, which also provide grasp input information for downstream tasks.
Motion speed is varied by commanding the robot to reach each waypoint in 4\,s (slow), 2\,s (medium), or 1--1.5\,s (fast).
In total, we collect 20 minutes of motion per DLO, 20 minutes per BDLO, and 15 minutes per cloth object, segmented into 5\,s intervals for a total of 1,320 motion sequences.

\subsubsection{Metrics}
\begin{figure*}[t]
    \centering
    \includegraphics[width=\linewidth]{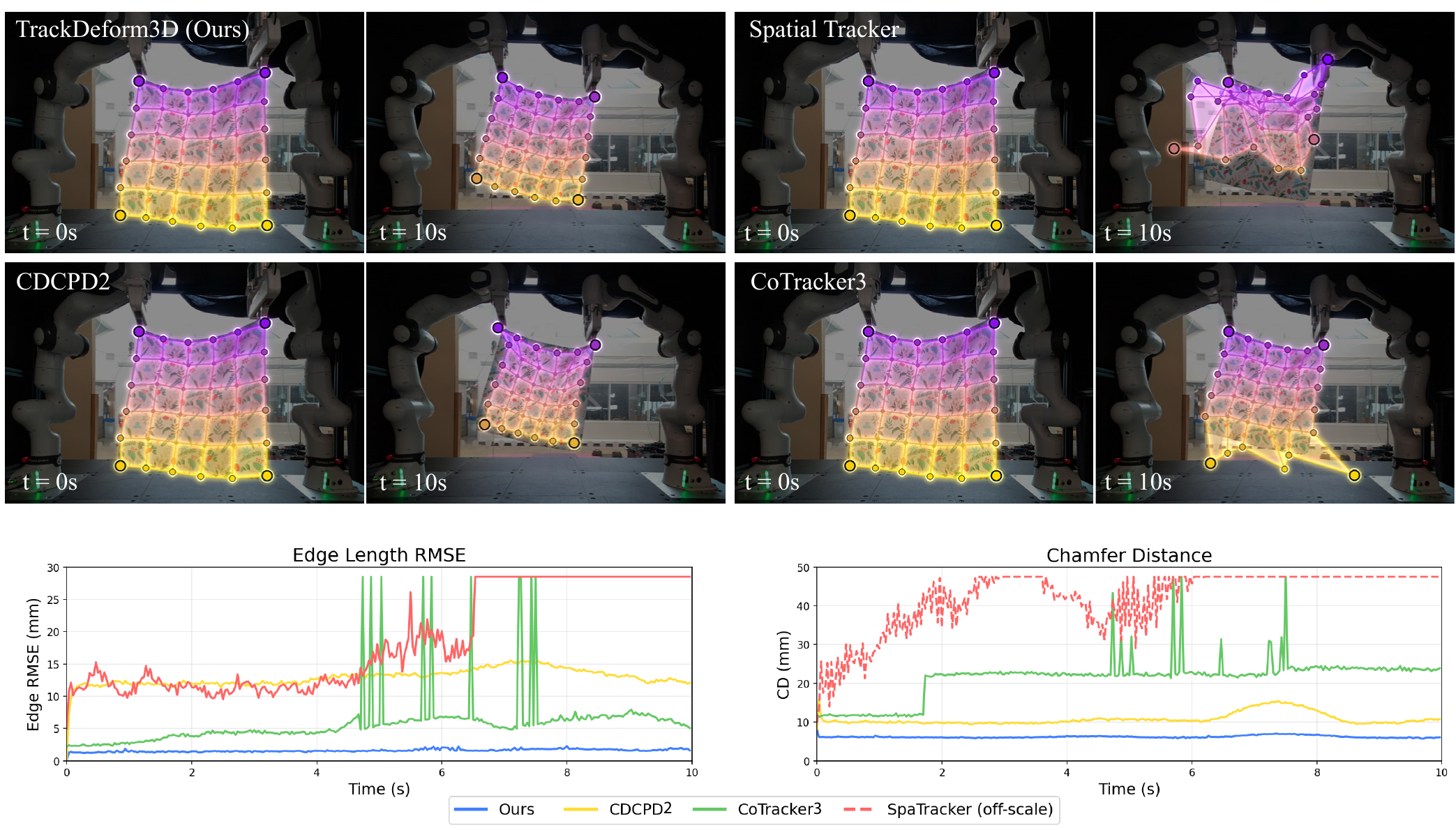}
    \caption{Upper: Qualitative comparison of keypoint tracking results between TrackDeform3D and all baseline methods on a cloth sequence at the start and end of manipulation. While CDCPD2, the best-performing baseline, maintains track of the deformable object, it exhibits inconsistent edge lengths and distorted mesh topology over time. SpatialTracker and CoTracker both show significant keypoint drift by the end of the sequence. Bottom: Edge length RMSE and Chamfer distance over time for all methods.}
    \label{fig:trackingper_vis_plot}
\end{figure*}

\begin{table*}[!t]
\centering
\setlength{\tabcolsep}{4pt}
\renewcommand{\arraystretch}{1.2}
\begin{tabular}{lcccc|cccc|cccc}
\toprule
\multirow{2}{*}{\textbf{Method}} &
\multicolumn{4}{c|}{\textbf{Rope}} &
\multicolumn{4}{c|}{\textbf{Branched Rope}} &
\multicolumn{4}{c}{\textbf{Fabric}} \\
\cmidrule(lr){2-5}\cmidrule(lr){6-9}\cmidrule(lr){10-13}
& \makecell{Abs.\\ (mm)} & \makecell{E\textless5\\(\%)} & \makecell{CD\\(mm)} & \makecell{F\textless10\\(\%)}
& \makecell{Abs.\\(mm)} & \makecell{E\textless5\\(\%)} & \makecell{CD\\(mm)} & \makecell{F\textless10\\(\%)}
& \makecell{Abs.\\(mm)} & \makecell{E\textless5\\(\%)} & \makecell{CD\\(mm)} & \makecell{F\textless10\\(\%)} \\
\midrule
CoTracker3~\cite{cotracker3}
& 42.31  & 1.0  & \cellcolor{second}8.54  & \cellcolor{best}\textbf{74.9}
& 20.35  & 24.7  & 42.87  & 36.4
& 25.09  & \cellcolor{second}61.3  & 20.60  & 13.2  \\
CDCPD2~\cite{pointcloud5}
& \cellcolor{second}3.89 & 25.3 & 10.73 & 63.7
& \cellcolor{second}5.08 & \cellcolor{second}42.8 & \cellcolor{second}8.02 & \cellcolor{second}80.3
& \cellcolor{second}13.28 & 20.4 & \cellcolor{second}13.97 & \cellcolor{second}47.7 \\
SpatialTrackerV2~\cite{spatialtrackerv2}
& 40.21  & \cellcolor{second}53.6  & 121.15  & 0.3
& 182.9  & 20.8  & 253.8  & 0.5
& 22.42  & 12.2  & 88.58  & 8.74  \\
\midrule
\textbf{TrackDeform3D (Ours)}
& \cellcolor{best}\textbf{3.19} & \cellcolor{best}\textbf{63.3} & \cellcolor{best}\textbf{8.04} & \cellcolor{second}74.1
& \cellcolor{best}\textbf{3.83} & \cellcolor{best}\textbf{87.8} & \cellcolor{best}\textbf{6.70} & \cellcolor{best}\textbf{87.3}
& \cellcolor{best}\textbf{2.99} & \cellcolor{best}\textbf{97.6} & \cellcolor{best}\textbf{7.52} & \cellcolor{best}\textbf{77.8} \\
\bottomrule
\end{tabular}
\caption{Baseline comparison. Abs. denotes absolute edge length error. CD denotes Chamfer distance. E\textless5 denotes the percentage of frames with Abs. below 5\,mm. F\textless10 denotes the fraction of frames with CD below 10\,mm over the entire trajectory.}
\label{tab:baseline_comparison}
\vspace{-4mm}
\end{table*}

We evaluate initialization consistency and tracking quality using three metrics.
Absolute edge length error (Abs.), reported as RMSE in millimeters, measures the deviation between distances of neighboring keypoints and their target edge lengths, reflecting how well the keypoint layout preserves the intended spacing.
Chamfer distance (CD), computed as the mean L2 distance between each predicted keypoint and its nearest neighbor on the segmented point cloud, reported in millimeters.
F-score complements these by jointly measuring precision and recall of predicted keypoints with respect to the observed point cloud, capturing how well keypoints both align with and cover the object geometry.

\subsubsection{Tracking Baseline Comparison}
We compare TrackDeform3D against three baselines under the same initialization.
CoTracker3~\cite{cotracker3} is a video-based point tracker that jointly tracks multiple points by exploiting long-range temporal correlations; its 2D tracking points are lifted to 3D using depth information.
CDCPD2~\cite{pointcloud5} is a point-cloud-based deformable object tracker that enforces geometric constraints to handle partial occlusions.
SpatialTrackerV2~\cite{spatialtrackerv2} is a recent 3D point tracking method that lifts 2D video tracking into 3D space.
Each method is applied to recorded RGB-D sequences comprising 1140\,s of rope, 1220\,s of branched rope, and 400\,s of cloth.
Quantitative results are reported in Table~\ref{tab:baseline_comparison}, and a qualitative comparison with per-frame error plots is shown in Fig.~\ref{fig:trackingper_vis_plot}.

Vision-based trackers~\cite{cotracker1, spatialtrackerv2} may fail due to noisy backgrounds, thin objects, or noisy and missing depth measurement.
CDCPD2 loses track during dynamic manipulation and inconsistently preserves edge lengths.
Additional visualizations are provided in the supplementary video.

\subsection{Ablation Studies}

\subsubsection{Tracking Ablation}
To evaluate the contribution of each constraint during tracking, we ablate the anchor and edge length components from~\eqref{eq:init_opt} across all three object categories, with results reported in Table~\ref{tab:track_ablation}.
For DLOs, the anchor constraint has no effect because the grasped endpoints already coincide with the end effector inputs, yielding results identical to the full formulation.
For BDLOs and cloth, removing the anchor constraint (\textit{w/o anchor}) causes keypoints to drift from their intended positions due to the increased degrees of freedom, resulting in higher absolute edge length error and lower E$<$5 scores.
Removing the edge length term (\textit{w/o repulsion}) leads to a substantial increase in absolute edge length error across all categories, as keypoints no longer maintain consistent spacing.
We omit the ablation on the point cloud projection step, as removing it would yield static keypoints that do not follow the observed geometry.
The full formulation consistently achieves the best performance across all metrics and object types, confirming that both constraints are essential for accurate and stable tracking.



\begin{table*}[!t]
\centering
\setlength{\tabcolsep}{4pt}
\renewcommand{\arraystretch}{1.0}

\begin{tabular}{lcccc|cccc|cccc}
\toprule
\multirow{2}{*}{\textbf{Method}} &
\multicolumn{4}{c|}{\textbf{Rope}} &
\multicolumn{4}{c|}{\textbf{Branched Rope}} &
\multicolumn{4}{c}{\textbf{Fabric}} \\
\cmidrule(lr){2-5}\cmidrule(lr){6-9}\cmidrule(lr){10-13}
& \makecell{Abs.\\(mm)} & \makecell{E\textless5\\(\%)} & \makecell{CD\\(mm)} & \makecell{F\textless10\\(\%)}
& \makecell{Abs.\\(mm)} & \makecell{E\textless5\\(\%)} & \makecell{CD\\(mm)} & \makecell{F\textless10\\(\%)}
& \makecell{Abs.\\(mm)} & \makecell{E\textless5\\(\%)} & \makecell{CD\\(mm)} & \makecell{F\textless10\\(\%)} \\
\midrule
w/o Anchor
& 3.19 & 63.3 & 8.04 & 74.1
& 9.56 & 62.7 & 15.00 & 60.0
& 6.44 & 56.7 & 22.72 & 58.0 \\
w/o Repulsion
& 74.34 & 6.8 & \cellcolor{second}14.49 & \cellcolor{second}59.9
& 52.35 & 9.9 & \cellcolor{second}8.43 & \cellcolor{second}77.6
& 48.21 & 70.0 & \cellcolor{second}10.46 & \cellcolor{second}65.4 \\
w/o Projection
& \cellcolor{best}\textbf{1.49} & \cellcolor{best}\textbf{96.0} & 120.83 & 3.9
& \cellcolor{second}3.51 & \cellcolor{best}\textbf{91.6} & 27.44 & 24.69
& \cellcolor{second}2.81 & \cellcolor{best}\textbf{99.6} & 17.11 & 44.5 \\
\midrule
\textbf{TrackDeform3D (Ours)}
& \cellcolor{second}3.19 & \cellcolor{second}63.3 & \cellcolor{best}\textbf{8.04} & \cellcolor{best}\textbf{74.1}
& \cellcolor{best}\textbf{3.83} & \cellcolor{second}87.8 & \cellcolor{best}\textbf{6.70} & \cellcolor{best}\textbf{87.3}
& \cellcolor{best}\textbf{2.99} & \cellcolor{second}97.6 & \cellcolor{best}\textbf{7.52} & \cellcolor{best}\textbf{77.8} \\
\bottomrule
\end{tabular}
\caption{Ablation study. Metrics are defined in Table~\ref{tab:baseline_comparison}.}
\label{tab:track_ablation}
\end{table*}

\subsubsection{Initialization Ablation}
\begin{figure}[t]
\centering
    \includegraphics[width=\linewidth]{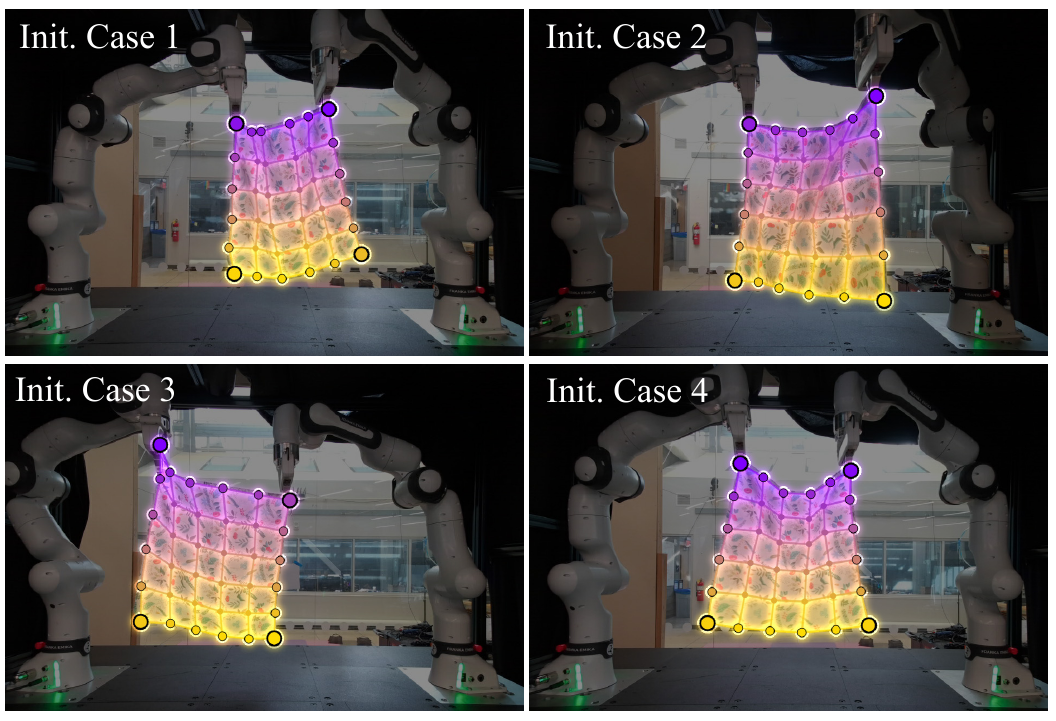}
    \caption{Demonstration of keypoint initialization consistency across four data collection sessions, showing that the full formulation produces stable and repeatable keypoint layouts.}
    \label{fig:initialization_ablation}
    \vspace{-5mm}
\end{figure}

To verify the importance of each component in~\eqref{eq:init_opt} for initialization, we ablate the anchor, edge length, and point cloud projection components on two representative object types: branched rope and fabric.
For each variant, initialization is re-run across multiple sessions and absolute edge length error is reported in Table~\ref{tab:init_consistency}.
The effects of removing the anchor and edge length constraints are consistent with those observed in the tracking ablation.
Removing point cloud projection causes keypoints to deviate from the observed object surface, resulting in geometrically invalid configurations.
The full formulation achieves the most stable and geometrically consistent initialization across sessions.
The initialization results across sessions is shown in Fig.~\ref{fig:initialization_ablation}.


\begin{table}[t]
\centering
\setlength{\tabcolsep}{6pt}
\renewcommand{\arraystretch}{1.0}
\begin{tabular}{lccc|ccc}
\toprule
\multirow{2}{*}{\textbf{Variant}} &
\multicolumn{3}{c|}{\textbf{Branched Rope}} &
\multicolumn{3}{c}{\textbf{Fabric}} \\
\cmidrule(lr){2-4}\cmidrule(lr){5-7}
& \makecell{Abs.\\(mm)} & \makecell{CD\\(mm)} & \makecell{F\textless10\\(\%)}
& \makecell{Abs.\\(mm)} & \makecell{CD\\(mm)} & \makecell{F\textless10\\(\%)} \\
\midrule
w/o Anchor
& 2.45  & 3.16  &95.3
&3.89  & 9.94  & 61.9  \\
w/o Repulsion
& 15.08  & \cellcolor{second}3.08  & \cellcolor{best}\textbf{96.4}
& 22.28  & \cellcolor{best}\textbf{5.50}  & \cellcolor{best}\textbf{86.2}  \\
w/o Projection
& \cellcolor{second}1.81  & 4.58  & 91.2
& \cellcolor{second}2.87  & 8.70  & 66.2  \\
\midrule
\textbf{Full}
& \cellcolor{best}\textbf{1.62} & \cellcolor{best}\textbf{3.06}  & \cellcolor{second}96.1
& \cellcolor{best}\textbf{2.58} & \cellcolor{second}6.66  & \cellcolor{second}79.0  \\
\bottomrule
\end{tabular}
\caption{Initialization consistency ablation. Metrics are defined in Table~\ref{tab:baseline_comparison}.}
\label{tab:init_consistency}
    \vspace{-3mm}
\end{table}

\subsection{Dataset Comparison}
We compare our dataset against two existing benchmarks: DOT~\cite{deformdatamocap2}, a large public dataset collected with textureless markers, and a motion capture dataset referred to as Mocap~\cite{deformdatamocap1}.
As shown in Table~\ref{tab:dataset_comparison}, our dataset provides broader object category coverage and significantly more motion sequences and total recording time.
Notably, it is the only dataset that includes branched deformable objects, which are important for applications such as wire harness modeling.
Mocap~\cite{deformdatamocap1} offers a richer variety of 2D object motion types and manipulation strategies, which we consider a valuable complement to our work.
In future iterations, we plan to expand the dataset with more diverse manipulation patterns and object categories.

\begin{figure}[t]
    \centering
    \includegraphics[width=\linewidth]{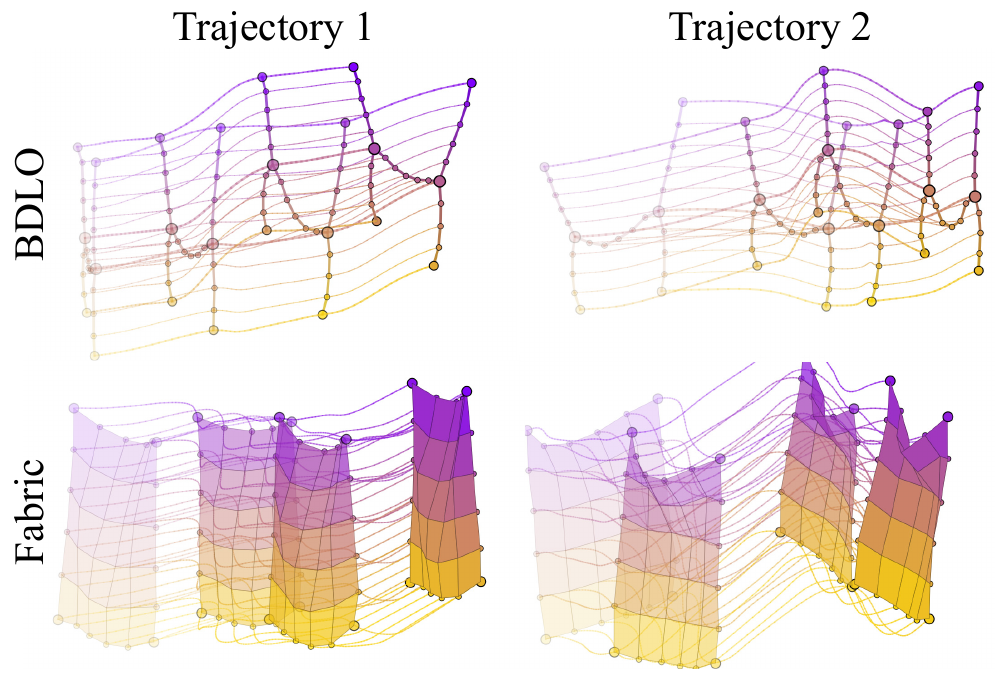}
    \caption{Example 3D trajectory visualizations from the collected dataset for a BDLO and cloth object. 
    Color encodes spatial position across keypoints, while transparency encodes temporal progression, with earlier frames appearing more transparent and later frames more opaque. 
    Each trajectory preserves the keypoint topology throughout the manipulation sequence.}
    \label{fig:trackingper_dataset_vis}
    \vspace{-6mm}
\end{figure}

The dataset will be released with 3D trajectories annotated with keypoint indexing and topology information.
Robot arm motion inputs will be accompanied by the indices of the grasped keypoints on the deformable object.
The dataset will be split into contact and non-contact subsets, with the workstation table mesh provided for contact-aware benchmarking.
A visualization of example trajectories is shown in Fig.~\ref{fig:trackingper_dataset_vis}.

\begin{table}[t]
\centering
\setlength{\tabcolsep}{5pt}
\renewcommand{\arraystretch}{1.0}
\begin{tabular}{lccc}
\toprule
 & \textbf{DOT}~\cite{deformdatamocap2} & \textbf{Mocap}~\cite{deformdatamocap1} & \textbf{Ours} \\
\midrule
DLO objects        & 1 & 0  & 2 \\
BDLO objects       & 0 & 0 & 2 \\
2D objects         & 2 & 4 & 2 \\
Total time         & 20 min & 9.8 min & 110 min \\
Motion sequences   & 220 & 120 & 1320 \\
\bottomrule
\end{tabular}
\caption{Dataset comparison with existing deformable object tracking benchmarks.}
\label{tab:dataset_comparison}
\vspace{-7mm}
\end{table}

\section{Conclusion}
We presented TrackDeform3D, a pipeline for robust 3D keypoint tracking of deformable objects during robotic manipulation.
By formulating both initialization and tracking as a unified constrained optimization, TrackDeform3D enforces anchor, edge length, and point cloud projection constraints to produce consistent and geometrically accurate keypoint trajectories across diverse object categories including DLOs, BDLOs, and cloth.
We also introduced a large-scale RGB-D dataset comprising over 1,300 motion sequences spanning multiple object types, stiffness levels, and manipulation speeds, annotated with 3D keypoint trajectories and topology information, which will be publicly released to support future research.
Experimental results demonstrate that TrackDeform3D outperforms existing baselines across all evaluated metrics and object categories, and ablation studies confirm the contribution of each optimization component.

\noindent\textbf{Limitations and Future Work:} The current approach assumes rectangular fabric geometry, a single-camera setup, and reliable anchor-point detection.
It may fail when anchor points are not properly identified, when object topology is complex or unknown, or when objects are occluded.
Future work will address these limitations.

\noindent\textbf{Acknowledgement:} The authors would like to gratefully thank reviewers for giving useful comments. 
This work is supported by the Ford Motor Company and the Air Force Office of Scientific Research under the
Award No: MURI FA9550-23-1-0400





\bibliographystyle{ieeetr}
\bibliography{references}
\end{document}